# Combine PPO with NES to Improve Exploration

**Li lianjiang**  LILIANJIANG@NEUQ.EDU.CN
*School of Control Engineering*
*Northeastern University at*
*Qinhuangdao*
*Qinhuangdao,Hebei 066004,CHN*

**Yang yunrong**  15032313387@163.COM
*School of Control Engineering*
*Northeastern University at*
*Qinhuangdao*
*Qinhuangdao,Hebei 066004,CHN*

**Li bingna**  LIBINGNA519@163.COM
*School of Control Engineering*
*Northeastern University at*
*Qinhuangdao*
*Qinhuangdao,Hebei 066004,CHN*

## ABSTRACT

We introduce two approaches for combining neural evolution strategy (NES) and proximal policy optimization (PPO): parameter transfer and parameter space noise. Parameter transfer is a PPO agent with parameters transferred from a NES agent. Parameter space noise is to directly add noise to the PPO agent's parameters. We demonstrate that PPO could benefit from both methods through experimental comparison on discrete action environments as well as continuous control tasks.

## 1. Introduction

Reinforcement learning is a branch of machine learning that studies how an agent can get more cumulative rewards when interacting with the environment. The agent must behave according to the previous experience but has to try actions that it has not selected in the past, which is called exploration-exploitation dilemma.[1] In the field of reinforcement learning, the trade-off is a long-term challenge. Frankly, the core of the balance between exploitation and exploration is how to get more exploration because the agent usually prefers to behave as the previous trajectory. An efficient exploration strategy could enable the agent to maximize the observation of the environment, ensuring that the behavior of the agent does not fall into a local optimum.

In order to make the exploration of agents more efficient, many methods have been proposed in the past few decades, but these methods are often based on complex additional structures, such as counting tables[2], density modeling in state space[3], learned dynamic models[3] or curiosity-driven exploration[4]. Other ways are to add additional time-related noise in the action space, such as the Policy gradient methods[5] and bootstrapped DQN[6]. But these methods usually lead to some unrealistic behaviors in high-dimensional or continuous tasks. Besides, abandoning the traditional Markov decision process (MDP) structure and innovatively applying the neural evolution strategy (NES) to the problem of reinforcement learning is also proposed[7]. NES can be regarded as a substitute for traditional MDP algorithms to some extent, but it is not sample efficient and requires a large amount of sample data. In addition to the above methods of increasing the exploration ability of agent, another approach is to add noise to the parameter space of policy[8][9][10]. This method is inspired by evolution strategy (ES) which solves problem by disturbing parameters of the neural network. The authors add noise to the parameter space of DQN[11], DDPG[12], TRPO[13], and compare them with algorithms adding noise in the action space. Experiments show that the former often performs better than the latter. PPO algorithm is a new algorithm proposed in 2017[14]. Due to its simple structure and ability to achieve complex discrete and continuous control tasks, PPO algorithm has become one of the most popular reinforcement learning algorithm. Although PPO algorithm performs well compared with previous algorithms, there are still shortcomings such as insufficient exploration, easy to fall into local optimum, and highly dependent on hyperparameter settings.

This paper mainly studies how neural evolution strategy can be effectively combined with PPO algorithm to improve exploration. Two approaches of combining NES and PPO are proposed, one is parameter migration and the other is parameter space noise. Experiments show that both methods can improve the exploratory behavior of PPO in continuous control tasks and discrete environments.

## 2. Background

This section mainly introduces the mathematical background of PPO, NES and noise neural network.

### 2.1 Proximal policy optimization

Proximal policy optimization (PPO) belongs to policy gradient methods and has an actor-critic[15] structure. It outperforms in high-dimensional environments and continuous control tasks. However, PPO still has the common problems of the MDP algorithm, such as not enough exploration and easy to fall into the local optimum. According to the paper[14], PPO is based on TRPO[13], which improves the objective function by adding a constraint, and this constraint ensures the extensive update of the objective. The improved objective function or actor has two forms:

$$L^{CLIP}(\theta) = \hat{E}_t[\min(\frac{\pi_\theta(a_t|s_t)}{\pi_{\theta_{old}}(a_t|s_t)}\hat{A}_t, clip(\frac{\pi_\theta(a_t|s_t)}{\pi_{\theta_{old}}(a_t|s_t)}, 1-\alpha, 1+\alpha)\hat{A}_t)] \quad (1)$$

And:

$$L^{KLPEN}(\theta) = \hat{E}_t[\frac{\pi_\theta(a_t|s_t)}{\pi_{\theta_{old}}(a_t|s_t)}\hat{A}_t - \beta KL[\pi_{\theta_{old}}(\cdot|s_t), \pi_\theta(\cdot|s_t)]] \quad (2)$$

Where, the form (1) is clipped surrogate objective method, $\pi_\theta$ is a policy, $\hat{A}_t$ is an estimator of the advantage function at timestep, and $\alpha$ is a hyperparameter, set to 0.2. Form (2) is adaptive KL penalty coefficient method, $\beta$ is the KL penalty coefficient.

For PPO, its critic has the following form:

$$L(\phi) = -\sum_{t=1}^{T}(\sum_{t'>t}\gamma^{t'-t}r_{t'} - V_\phi(s_t))^2 = -\sum_{t=1}^{T}\hat{A}_t^2 \quad (3)$$

As for the clipped surrogate objective method, the goal of actor is to maximize $L^{CLIP}(\theta)$, and the critic aims to minimize $\hat{A}_t$. Actor would adapt the new policy based on the old policy according to $\hat{A}_t$, the larger $\hat{A}_t$ is, the more likely new policy occurs. And we add a clip item to the objective in order to prevent the policy from updating excessively. In terms of adaptive KL penalty coefficient method, this approach restricts the update extent of policy with the similarity of new policy and old policy. The standard is as follows:

If $KL[\pi_{old}|\pi_\theta] > 1.5KL_{target}$, $\beta \leftarrow 2\beta$

If $KL[\pi_{old}|\pi_\theta] < KL_{target}/1.5$, $\beta \leftarrow \beta/2$

In this paper, we choose the clipped surrogate objective method to build PPO, for this method is better than the other.

**2.2 Neural evolution strategy**

In reinforcement learning, neural evolution strategy (NES)[7] applies Gaussian noise to the parameters of the neural network, and uses the perturbed neural network to interact with the environment to collect rewards, thus enabling the agent to explore the environment. Compared with algorithms that increase noise in action space, such as the policy gradient methods, this kind of exploration can enhance a more diverse exploration behaviors, thus improving the exploration performance of agent.

The simple single-threaded NES algorithm is as follows[7]:

---
**Algorithm 1 Eovolution Strategies**
1: **Input: Learning rate** $\alpha$ **, noise standard deviation** $\sigma$ **, initial policy parameters** $\theta_0$

2: **for** $t = 0,1,2...$ **do**

3:     **Sample** $\varepsilon_1,...\varepsilon_n \sim N(0,I)$

4:     **Comupte returns** $F_i = F(\theta_t + \sigma\varepsilon_i)$ **for** $i = 1,...,n$

5:     **Set** $\theta_{t+1} \leftarrow \theta_t + \alpha \dfrac{1}{n\sigma} \sum_{i=1}^{n} F_i \varepsilon_i$

6: **end for**

---

The NES algorithm can also achieve multi-core parallel work well, which can reduce training time and improve training efficiency. The parallelized version of the NES algorithm is as follows[7]:

---
**Algorithm 2 Parallelized Eovolution Strategies**
1: **Input: Learning rate** $\alpha$ **, noise standard deviation** $\sigma$ **, initial policy parameters** $\theta_0$

2: **Initialize:** $n$ **workers with known random seeds, and initial parameters** $\theta_0$

3: **for** $t = 0,1,2...$ **do**

4:     **for each worker** $i = 1,...n$ **do**

5:         **Sample** $\varepsilon_i \sim N(0,I)$

6:         **Comupte returns** $F_i = F(\theta_t + \sigma\varepsilon_i)$

7:     **end for**

8:     **Send all scalar returns $F_i$ from each worker to every other worker**

9:     **for each worker $i = 1,...,n$ do**

10:       **Reconstruct all perturbations $\varepsilon_j$ for $j = 1,...,n$ using known random seeds**

11:       **Set $\theta_{t+1} \leftarrow \theta_t + \alpha \dfrac{1}{n\sigma} \sum_{j=1}^{n} F_j \varepsilon_j$**

12:   **end for**

6: **end for**

---

The exploration principle of the NES algorithm is different from the PPO algorithm of the MDP structure. It does not perturb action space to explore, but perturb the parameter space. Therefore, combining the NES algorithm with the PPO algorithm can theoretically double the exploration and inspire the agent interaction with the environment.

### 2.3 Noisy neural network

According to[9], noisy neural network is a type of network whose internal parameters, that is, weights and biases are disturbed during training. In PPO, the policy $\pi_\theta$ is represented by the actor's neural network, then the parameters $\theta$ are the internal parameters of the neural network $w$ and $b$. Let Gaussian noise $\varepsilon$ obey the standard normal distribution, that is, $\varepsilon \sim N(0,1)$ and $\theta = \mu + \sigma \times \varepsilon$ denotes the perturbed parameters, where $\mu$ and $\sigma$ both are a set of vectors of learnable parameters.

Let $x$ denotes the inputs of the actor neural network, and $y$ denotes the outputs, then a linear layer of a neural network expression is:

$$y = wx + b \tag{4}$$

The neural network after adding noise is:

$$y = (\mu_w + \sigma_w \times \varepsilon_w)x + (\mu_b + \sigma_b \times \varepsilon_b) \tag{5}$$

The graphical representation of noisy neural network is shown in Figure 1:

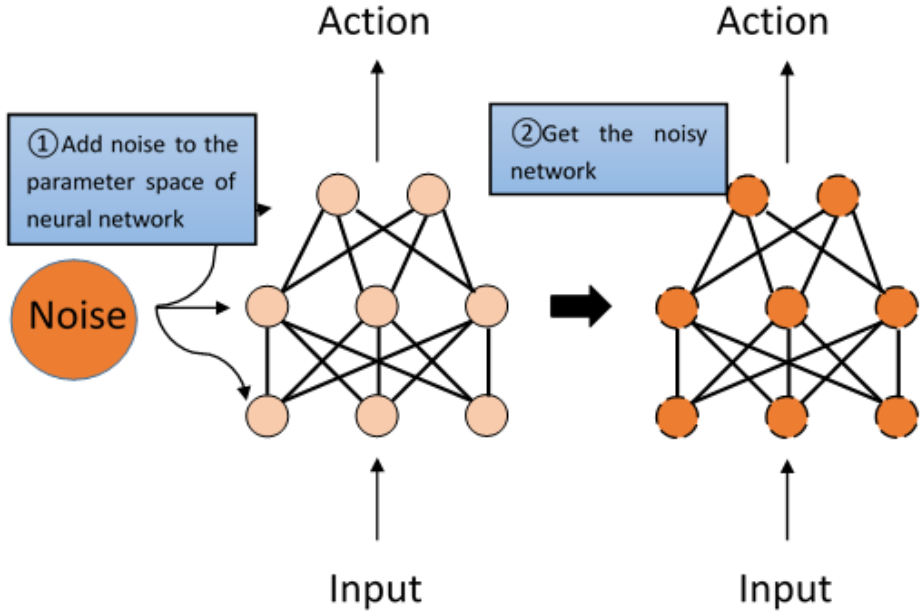

Figure 1: The process of adding parameter space noise.

For the Gaussian noise addition type, we tried two approaches[9], one is independent Gaussian noise and the other is factorized Gaussian noise.

### a. Independent Gaussian noise

For independent Gaussian noise, the noise of $w$ and $b$ is independent of each other and comes from independent Gaussian distributions. With $n$ input and $m$ output, the number of noise constants is $nm+m$.

### b. Factorized Gaussian noise

For factorized Gaussian noise, first, we factorize $\varepsilon_w$:

$$\varepsilon_w = f(\varepsilon_i)f(\varepsilon_j) \tag{6}$$

Where, $i=0,1,..(n-1)$, $j=0,1,2..(m-1)$, $f(x)=\text{sgn}(x)\sqrt{|x|}$

$$\varepsilon_b = f(\varepsilon_j) \tag{7}$$

The number of Gaussian noise constants generated by this method is $n+2m$, and is reduced compared to the previous independent Gaussian noise, thereby this kind of noise could save computation time.

### 2.4 Transfer learning

As an important branch of machine learning, transfer learning focuses on transferring learned knowledge to a brand new problems. Transfer learning is also a learning process, which utilizes the direct similarity of data, tasks or models to inject learned knowledge in old domain to new domain. Generally, let $D_S$ be source domain[16], $T_S$ be learning task, $D_T$ be target domain, $T_T$ be target task, transfer learning aims to help improve the learning of the target predictive function $f_T(\bullet)$ in $D_T$ using the knowledge in $D_S$ and $T_S$, where $D_S \neq D_T$, $T_S \neq T_T$.

Transfer learning can improve learning in three aspects: firstly, the initial performance before completing the learning is better than the without transfer; second, the time spent completing the learning is shorter than learning from scratch; third is the final learning performance is better than the level without transfer. Figure 2 illustrates these three aspects:

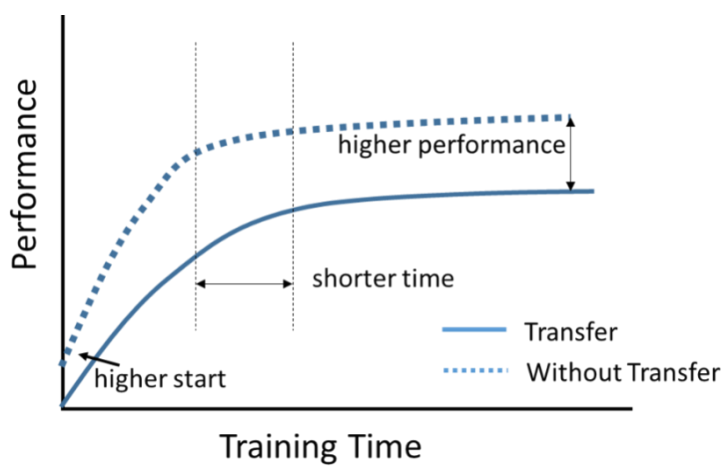

Figure 2: Three aspects transfer learning may improve learning.

According to the authoritative review article[16], the basic methods of transfer learning can be divided into four classes. The four basic approaches are: sample-based transfer, model-based transfer, feature-based transfer, and relationship-based transfer. And the application of transfer learning is not limited to special fields, but to all applications that meet the conditions of the above four methods. Transfer learning in neural networks has proved to be effective at enhance network ability[17]. Initializing a network with transferred features from almost any number of layers can produce a boost to performance and generalization. Reinforcement learning is developing with neural networks closely, such as deep reinforcement learning combines reinforcement learning with deep neural network. In theory, neural network transfer can update the performance of agent in deep reinforcement learning[18].

## 3. Improvement

This section describes two methods for improving PPO with NES: parameter transfer and parameter space noise.

### 3.1 Parameter transfer

According to the theory of transfer learning[16], in deep learning, parts or all of the parameters of an already trained neural network can be migrated to a completely blank neural network, instead of initializing neural network parameters from scratch, which could shorten the training time and even improve accuracy. The premise is that the application scenarios of the two neural networks should be similar. For example, the neural network that recognizes the cat can use the migration learning to identify the dog.

The graphical representation of parameter transfer is shown in Figure 3:

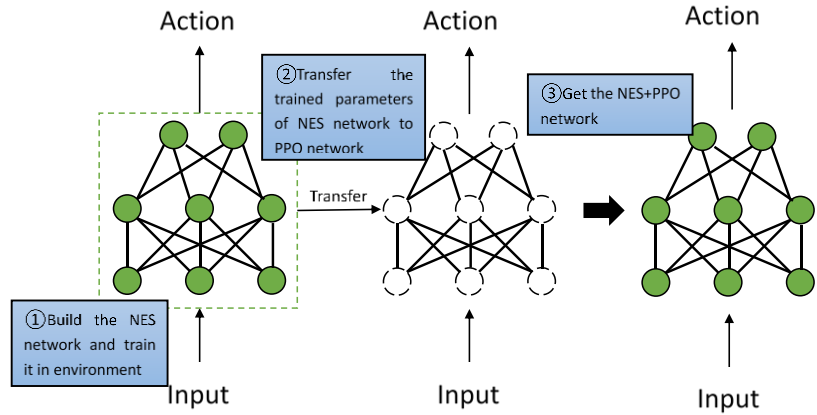

Figure 3: Parameter transfer has three steps: build the NES network and train it in the environment; transfer the trained parameters of NES network to PPO network of policy; get the NES+PPO network.

Along the same lines, we attempt to apply transfer learning to reinforcement learning problems. The steps of parameter transfer method are as follows:

a) Run a reinforcement learning environment with parallelized NES.
b) Save the parameters of NES neural network with good performance.
c) Transfer the parameters of NES neural network to the policy network of PPO.
d) Run the same reinforcement learning environment with NES+PPO.

Let $\theta_0$ denote the initial parameters of the neural network in NES, and

$\theta_{NES}$ denote the parameters after training, then after parameter transfer, the parameters of PPO network initialized to $\theta_{PPO} = \theta_{NES}$.

After parameter transfer, the policy network parameters of PPO are not randomly initialized from the beginning, but are initialized to parameters with good performance, so that PPO can update on the shoulders of giant. In theory, this approach can save time of parameter initialization, enhance sample efficiency and improve the performance of PPO.

### 3.2 Parameter space noise

#### 3.2.1 INITIALIZATION OF NOISY NEURAL NETWORK

In this paper, the above two Gaussian noises are used to perturb the policy network of PPO. In the case of independent Gaussian noise, the initial value of $\mu$ is randomly sampled from the uniform distribution $\left[-\sqrt{\frac{3}{n}}, +\sqrt{\frac{3}{n}}\right]$, and the initial value of $\sigma$ is set to 0.0017. For factorized Gaussian noise, random samples $\mu$ are taken from the uniform distribution interval $\left[-\sqrt{\frac{1}{n}}, +\sqrt{\frac{1}{n}}\right]$, and the initial value of $\sigma$ is set to $\frac{\sigma_0}{\sqrt{n}}$, where $\sigma_0 = 0.5$.

#### 3.2.2 NOISYNET-PPO

The policy of PPO is represented by actor, so we add noise to the fully connected layer parameters of actor, and critic does not be changed any more. Then the parameter $\theta$ is about to be modified as $\mu + \sigma \times \varepsilon$. The loss function of the modified PPO becomes:

$$L^{CLIP}(\mu, \sigma) = \hat{E}_t[\min(\frac{\pi_{\mu,\sigma}(a_t | s_t)}{\pi_{\mu_{old},\sigma_{old}}(a_t | s_t)}\hat{A}_t, clip(\frac{\pi_{\mu,\sigma}(a_t | s_t)}{\pi_{\mu_{old},\sigma_{old}}(a_t | s_t)}, 1-\alpha, 1+\alpha)\hat{A}_t)] \quad (8)$$

## 4. Experiments

This section focuses on the following two issues in continuous control tasks and discrete environments:

a. Can the combination of NES and the PPO improve the exploration ability?
b. Does PPO combined with the NES algorithm become more stable?

### 4.1 Continues control tasks

**Parameter space noise** In order to study whether parameter space noise can improve the exploration of PPO in continuous control tasks, we choose Unity[19] as the experimental platform, and build an experiment environment named "RollerBall" under this platform. The environment consisted of a floor, a ball, a square, and a camera above the environment to overlook the audience. The ball is regarded as the agent, the square is regarded as the target, and the ball could roll on the floor at a constant speed. The goal of the agent is to hit the square without dropping the floor. During the training process, each time the ball successfully completes a collision, it will receive corresponding rewards. If the collision fails, the agent will be punished. If the agent drops from the floor, it will be punished more. At the end of each episode, the position of the square will change randomly, which keeps the ball from repeating to move to a single direction. The experiment flow chart is shown in Figure 4:

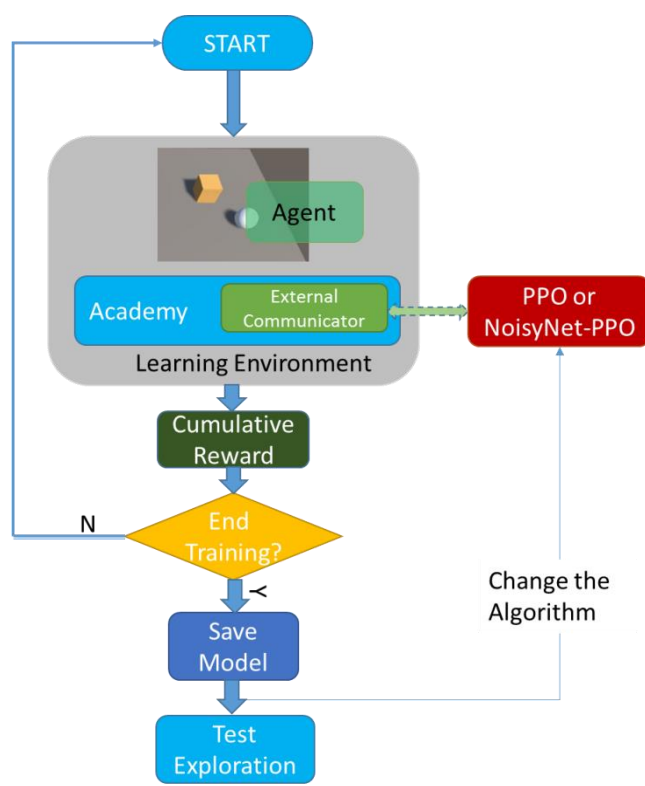

Figure 4: The flow chart of the "RollerBall" environment.

Compared to the baseline of PPO, we run NoisyNet-PPO with independent Gaussian noise and factorized Gaussian noise in the above environment respectively. The maximum number of training rounds is set to 500,000 times. The basic setting of the PPO algorithm is that the actor and critic neural networks contain two hidden layers in addition to the input and output layers, and the number of neurons in each layer is 128.

In order to further test the performance of the trained model in the environment, we reload the three models that have been trained in the

"RollerBall" environment, and sets the following scoring criteria: the ball hits 1 point for each successful collision, and will not score if the impact fails. If the ball drops the floor, it will lose 10 points. The cumulative score limit is set to 3000 points. When the cumulative score reaches the upper limit, the test is stopped and the test time is recorded. The environment diagram during the experiment is in Figure 5:

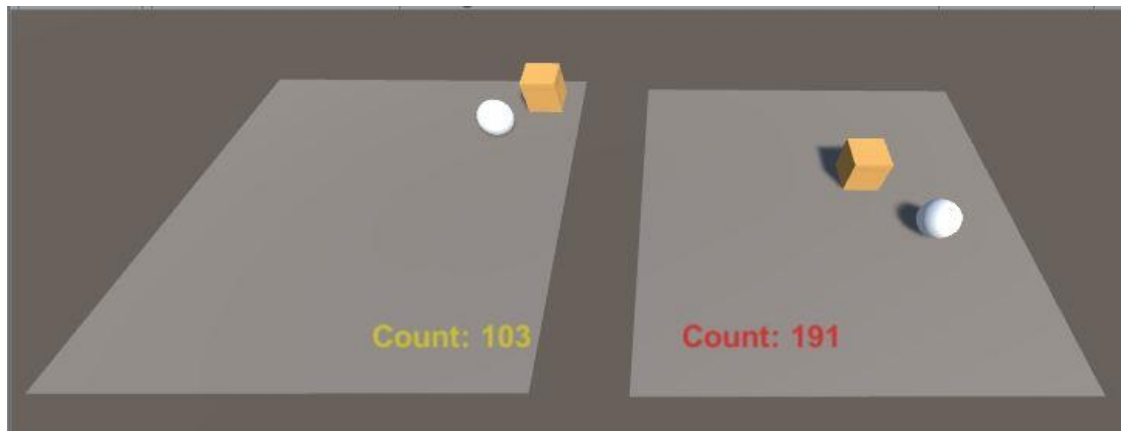

Figure 5: The testing environment of PPO and NES+PPO, the left is tested with PPO and the right is tested with NoisyNet-PPO. And we set a board to show the socres in the testing proceeding.

After the test is completed, the score curve of the ball is shown in Figure 6:

As is shown in Figure 6, NoisyNet-PPO spend a shorter time than the original PPO in terms of the time to reach the limit score, and the performance of NoisyNet-PPO with factorized Gaussian noise is much better than the independent Gaussian noise. In addition, the curves of PPO and NoisyNet-PPO with independent Gaussian noise show a distinct "small platform", that is, the scores remain unchanged as time goes by. In fact, in the real test environment, the ball hesitates near the square, but does not hit the square. This shows that the agent is in a period of stagnation during this time and does not explore the environment. The curve of NoisyNet-PPO with factorized Gaussian noise + PPO has few obvious "small platforms", which indicates that in continuous control task, factorized Gaussian noise is more helpful to improve the exploration of PPO.

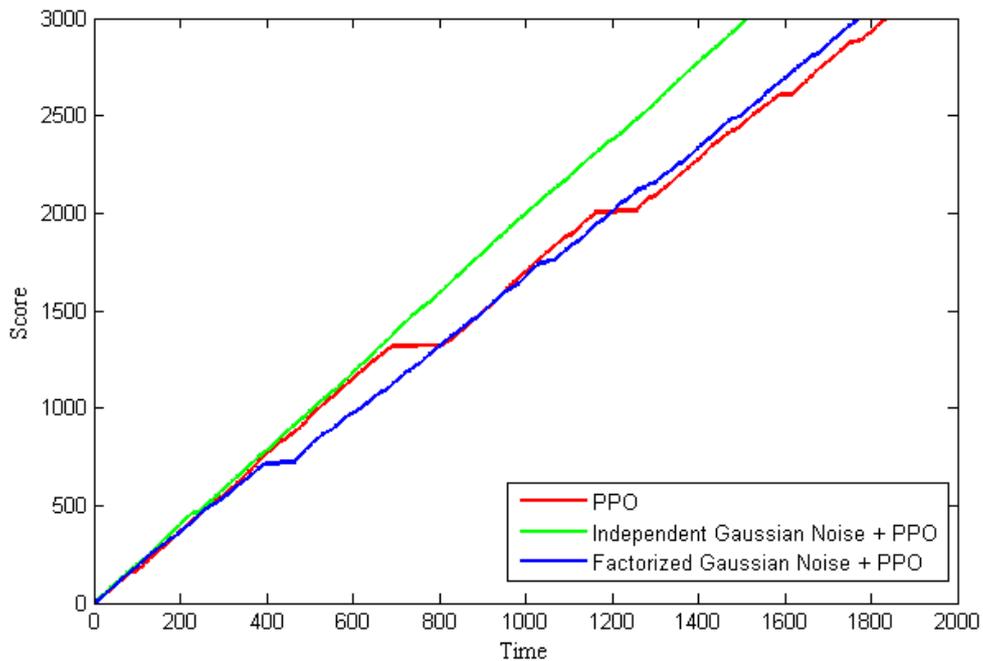

Figure 6: The score curve of the ball with three settings respectively. We could see NoisyNet-PPO performs better than PPO.

**Parameter Transfer** This section uses ***Pendulum*** in OpenAI Gym[20] as the training environment. Comparing the original PPO with NES+PPO after parameter transfer, we can observe the performance of the two algorithms in the simple continuous control task.

The figure below shows that in **Pendulum**, the final cumulative reward of the NES+PPO is 200 more than the cumulative reward of the original PPO, which indicates that parameter transfer has played a significant role in improving PPO exploration.

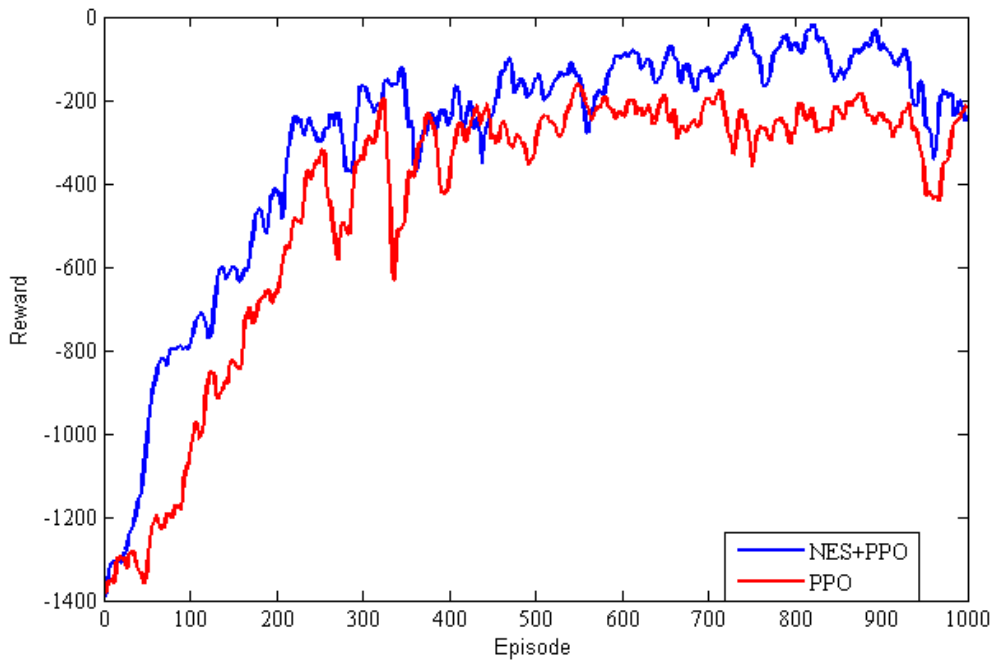

Figure 7 Comparison of PPO and NES+PPO

In the above experiment, the hyperparameter $\varepsilon$ of PPO is default to the optimal value $0.2$. Therefore, in order to explore whether the performance of algorithms will change when the hyperparameter is not optimal, the value is changed to 0.1 and 0.3, respectively. The experimental results are shown in Figure 7 below. The following are the effect diagrams of $\varepsilon = 0.1$ and $\varepsilon = 0.3$:

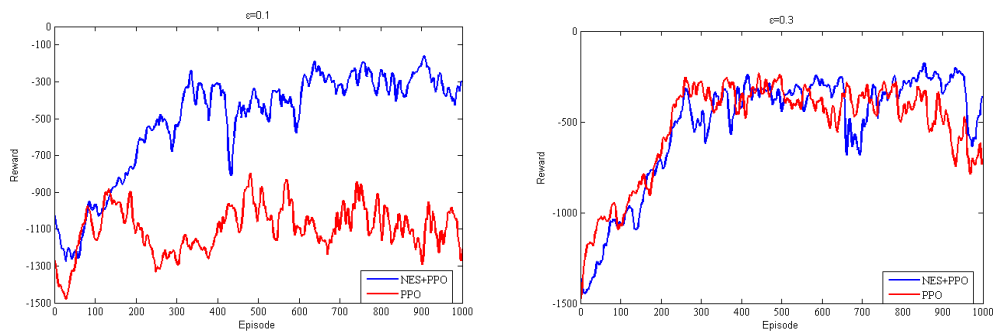

Figure 8 Comparison of NES+PPO and PPO after modifying PPO's hyperparameter

The following figure compare NES+PPO with NES when the hyperparameter $\sigma$ of NES is changed from 0.1 to 0.01 and 0.06, respectively.

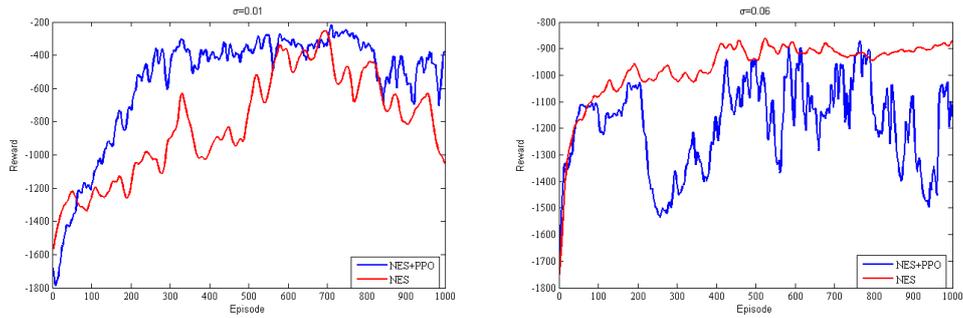

Figure 9 Comparison of NES+PPO and PPO after modifying NES's hyperparameter

It can be seen from the above pictures that PPO and NES are sensitive to hyperparameter changes, and their performance become worse. However, the performance of NES+PPO is relatively stable in most cases. This shows that in the continuous control task, parameter transfer can improve the stability of the original PPO.

### 4.2 Discrete environment experiment

**Parameter space noise** The experimental environment is a square plane surrounded by walls built under the Unity experimental platform. The target object is a cube brick that moves around the agent in a circular motion. Its moving radius is a random integer between 3 and 19, moving speed is divided into three types: *static*, *slow*, and *fast*, and moving direction is either clockwise or counterclockwise. The position of the agent is fixed in the middle of the plane, and its observation includes sensible ray, move direction of the target, and whether the agent is shooting.

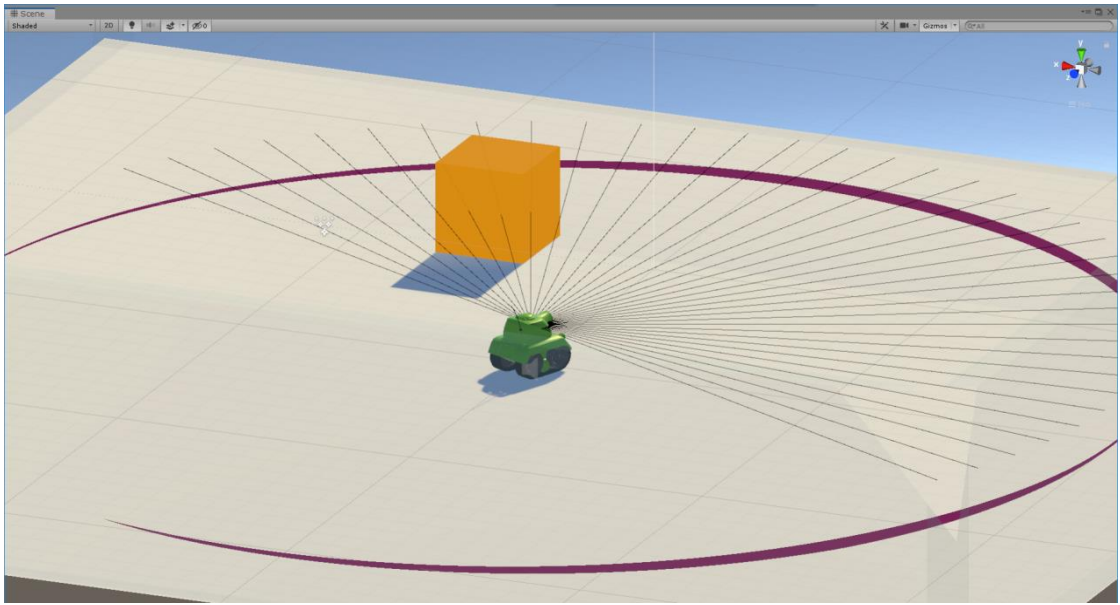

Figure 10: The discrete environment with a tank, a target and rays.

As is shown in Figure 10, the ray is a set of fixed length segments, which are centered at the position where the agent is located and are emitted outward at a certain angle. The perceptual ray can penetrate all objects present in the length range and inform the agent "what objects exist at an angle" through the function; at the same time, it can return the relative distance value between the first object and the agent. The effect of the sensible ray is similar to a laser radar, which gives the agent the ability to observe the environment.

In this experiment, the length of each sensible ray is 20, the angle of the ray ranges from $0°$ to $180°$, and the middle interval is $5°$—that is, the agent has a total of 37 sensible rays, and the range of perceived angles covers the entire range. Positive round half a week. Since the agent does not need to sense the wall at this time, the sensible ray only needs to recognize the orange brick. The movement of the agent is discrete, including left turn, right turn, still, fire. The agent has three brains: Static Brain, Slow Brain, and Fast Brain, which respectively deal with the target moving speed as static, slow, and fast—where the speed of the object corresponding to "stationary" is 0; the speed of the object corresponding to "slow" is 0.5; the speed of the object corresponding to "fast" is 1.The reward mechanism of the agent is as follows: if the agent hits the target during the training time, the reward value is +1 ; if the target is not hit, the reward value is -0.5; in addition, the agent will get a time penalty of -0.0005 every time spend a time step.

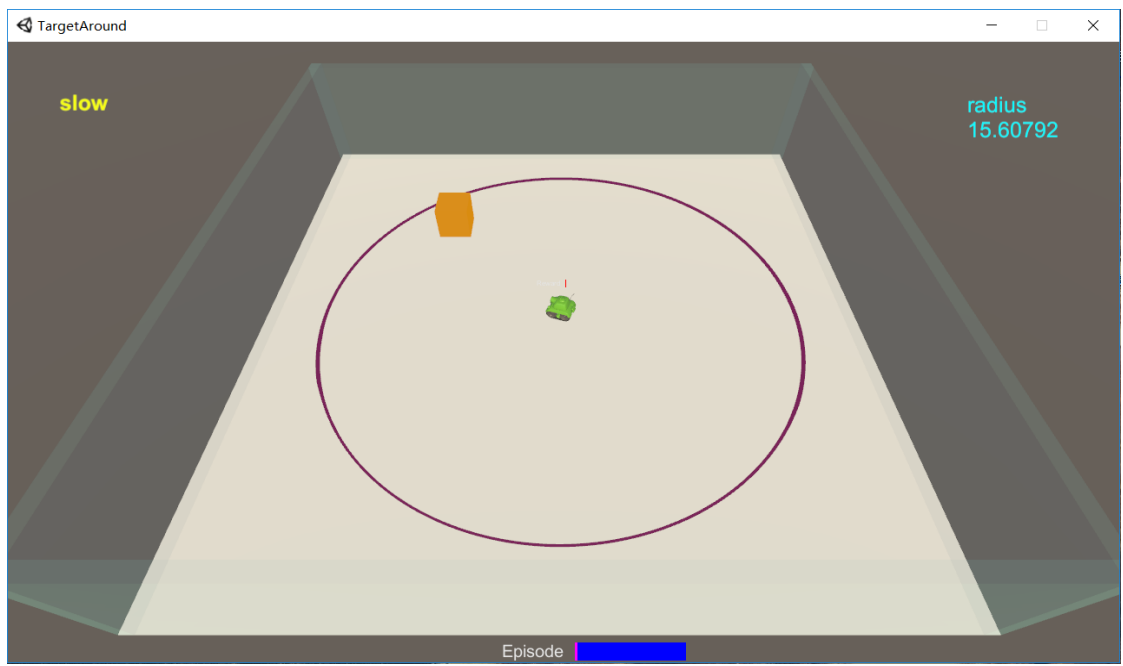

Figure 11: The training process diagram. The target is at a low speed.

In this environment configuration, we use PPO and NoisyPPO(factorized

Gaussian noise) to train 5 million times each. The training curves are as follows:

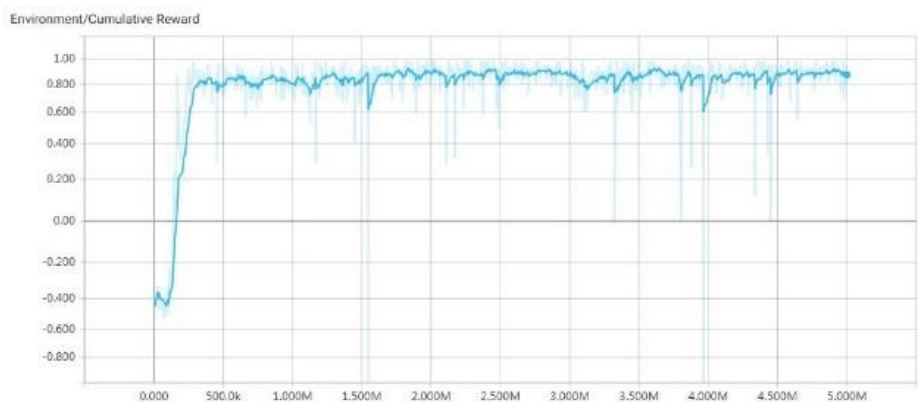

Figure 12: The agent is trained with PPO and the target is static.

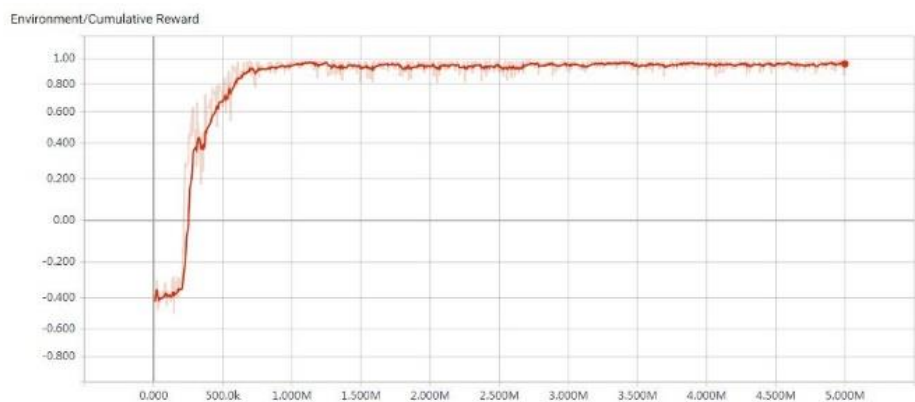

Figure 13: The agent is trained with PPO and the target is at a slow speed.

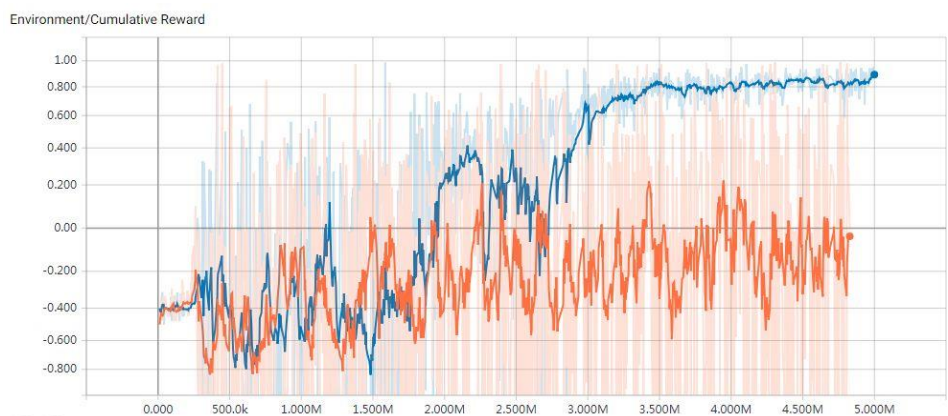

Figure 14: The agent is trained with PPO(red curve) and NoisyNet-PPO(blue curve) respectively and the target is at a fast speed.

When the target is at a **static speed** and **slow speed**, PPO achieves a good training effect, and the agent can pick up hitting the target. However, when the target speed is **fast**, the reward curve of the PPO algorithm rises to 0 and then does not continue to rise. This indicates that the agent is in local optimum and will not continue to explore, not to say hit the target. After adding the parameter space noise, the reward curve of the PPO algorithm starts along a similar trajectory as former, but at the middle of the training process, the curve begins to rise and reaches equilibrium in the second half of the training process. This shows that PPO jumps out of the local optimum and continues to explore the environment with the help of parameter space noise. The experiment results indicate that in the discrete environment, increasing the parameter space noise is still applicable to the PPO algorithm, which can prevent the PPO algorithm from converging prematurely to a local optimum and ensure the exploration ability of PPO.

### 5. Conclusion

We have presented two methods based on NES for PPO exploration in deep reinforcement learning that show significant improvements both in continuous control tasks and discrete environment. The methods are named as parameter transfer and parameter space noise. Parameter transfer refers to transferring the trained neural network parameters of NES to PPO; parameter space noise is to add noise to the internal parameters of the PPO's policy. The experiments of the continuous control tasks show that both approaches can improve the exploration performance of the PPO algorithm to different degrees and optimize the performance of the agent. Parameter transfer can also alleviate the dependence of PPO on hyperparameters, and improve the stability of PPO to some extent without deliberately adjusting hyperparameters. The results of discrete environment show that parameter space noise can effectively prevent the PPO algorithm from falling into local optimum, improve the exploration ability of the PPO algorithm, and achieve the goal that PPO failed to achieve in the past.